\documentclass[sn-basic]{sn-jnl}

\usepackage{natbib}
\usepackage{CJKutf8}

\jyear{2021}%

\theoremstyle{thmstyleone}%
\theoremstyle{thmstyletwo}%

\theoremstyle{thmstylethree}%

\begin{document}

\title[A study of conceptual language similarity: comparison and evaluation]{A study of conceptual language similarity: comparison and evaluation}


\author*[1,2]{\fnm{Haotian} \sur{Ye}}\email{yehao@cis.lmu.de}

\author[1,2]{\fnm{Yihong} \sur{Liu}}\email{yihong@cis.lmu.de}

\author[1,2]{\fnm{Hinrich} \sur{Schütze}}\email{inquiries@cislmu.org}

\affil[1]{\orgdiv{Center for Information and Language Processing},
                \orgname{LMU Munich},
                \country{Germany}}

\affil[2]{
    \orgname{Munich Center for Machine Learning (MCML)},
    \country{Germany}}


\abstract{An interesting line of research in natural language processing (NLP) aims to incorporate linguistic typology to bridge linguistic diversity and assist the research of low-resource languages. While most works construct linguistic similarity measures based on lexical or typological features, such as word order and verbal inflection, recent work has introduced a novel approach to defining language similarity based on how they represent basic concepts, which is complementary to existing similarity measures. In this work, we study the conceptual similarity in detail and evaluate it extensively on a binary classification task.}




\maketitle

\section{Introduction}\label{sec:intro}

There are over 7000 languages in the world today, which are categorized into more than 400 language families \citep{joshi-etal-2020-state, harald_hammarstrom_2022_7398962}. The increasing availability of unlabeled data in electronic form has contributed to significant progress in the development of multilingual NLP, a prominent example of which is the recent surge of multilingual pre-trained language models \citep{devlin-etal-2019-bert, conneau-etal-2020-unsupervised, xue-etal-2021-mt5}. However, such progress has so far excluded the great majority of the world's languages, most of which are low-resource. For these languages, data scarcity remains an issue to be solved. A few approaches are motivated to leverage linguistic information from high-resource languages (e.g. English and Chinese) to benefit languages that are less well-studied. Language similarity often plays an important role in this process. It has been shown that similar languages can aid the performance of transfer learning \citep{kim-etal-2017-cross, ahmad-etal-2019-difficulties, lauscher-etal-2020-zero} and joint learning \citep{cohen-etal-2011-unsupervised, navigli-ponzetto-2012-joining, wang-etal-2021-contrastive}.

Existing typologies typically categorize languages according to their geographical (e.g. the continent where the language is spoken), phylogenetic (genealogical relationships of languages), or structural (e.g. syntax and grammar) similarities.
The main source of typological information remains manually constructed databases of typological features, such as Glottolog \citep{harald_hammarstrom_2022_7398962}, PHOIBLE \citep{moran2014phoible}, WALS \citep{wals}, and the more recent Grambank \citep{skirgaard2023grambank}, although some approaches attempt to learn such features automatically where the databases' coverage is inadequate, e.g. through word alignment \citep{mayer-cysouw-2012-language, ostling-2015-word}.

Contrary to language similarity based on typological features, we study conceptual language similarity as the topic of our work. Conceptual language similarity is introduced in our previous work \citep{liu2023conceptualizer}, in which we propose the Conceptualizer, which uses a two-step pipeline to align basic concepts across 1335 languages with the help of a superparallel dataset: the Parallel Bible Corpus (PBC) \citep{mayer-cysouw-2014-creating}. Unlike most existing approaches, the Conceptualizer aims to find similarities and differences in how languages divide the world into concepts and what they associate with them. For instance, Chinese, Japanese, and Korean all associate the ``mouth'' concept with ``entrance'' due to influence from the Chinese character ``\begin{CJK*}{UTF8}{bsmi}口\end{CJK*}''. However, this association between ``mouth'' and ``entrance'' is missing from European languages, an indication that the three East Asian languages share a similar conceptualization that diverges from European languages with respect to the ``mouth'' concept. Based on the belief that conceptualizations of a language reflect the thoughts of its speakers \citep{deutscher2010through}, conceptual similarity represents a novel perspective of viewing the relatedness of languages, which is complementary to conventional measures based on lexical and typological similarities.

In this work, we extend our previous work by extensively evaluating and comparing conceptual language similarity to other similarity measures.
Following the original work, we perform the evaluation on a binary classification task where we predict whether most of a language's neighbors belong to the same language family. We also compare the results against those achieved using existing similarity measures.
To the best of our knowledge, no prior work has carried out an empirical evaluation of different language representations for predicting genealogical language similarity.
We show that conceptual similarity is weaker than but complementary to similarities based on lexical and typological features.

\section{Related Work}

There is a large body of work on language similarity. Most of the existing approaches use lexical or typological features of the respective languages. We present a few categories of language similarity measures in this section.

\subsection*{Lexical similarity}

Lexical similarity is a surface similarity measure and can be used to, e.g., determine whether two language variants are likely dialects. For example, Ethnologue considers a variant with a lexical similarity of $\gt$ 85\% a potential dialect \citep{eberhard2019ethnologue}.
One method of measuring it is comparing a multilingual lexicostatistical list such as the PanLex Swadesh list \citep{kamholz2014panlex}, which contains 100 words describing basic concepts in over 2000 languages (207 words in the extended version) \citep{swadesh2017origin}.
Larger lists, such as NorthEuraLex \citep{dellert2020northeuralex}, which contains 1016 concepts in 107 languages, are also used, e.g. in a study by \citet{rama2020probing}. The language coverage, however, is much lower than the PanLex Swadesh list.
\citet{holman2008explorations} show that the Swadesh-100 list can be reduced to a shorter list consisting of the 40 most stable elements while increasing the accuracy of language classification. The shortened concept list is incorporated in the ASJP database, which contains the list in 5590 languages \citep{soren2018asjp}. ASJP word lists are used by \citet{ostling2023language} to evaluate lexical distances between 1012 languages. Their method uses the mean normalized Levenshtein distance between each pair of concepts.
Alternatively, the pairwise Levenshtein distance can be substituted by a simple longest common substring method that effectively also measures the amount of shared lexical information between two languages.

\subsection*{Genealogical similarity}

The genealogical similarity between two languages is measured based on a genealogical, or phylogenetic, language tree. The most straightforward form of genealogical similarity is a binary indicator of whether two languages belong to the same top-level family, e.g. Indo-European or Sino-Tibetan (1 for the same family, 0 for different families). We can make the metric more sophisticated by introducing intermediate levels of the language tree. Below are the complete paths for two languages, Hungarian (hun) and Estonian (ekk), which include all tree levels (data from Glottolog).



{
\vspace{3mm}
\footnotesize{
    hun: $\text{Uralic} \rightarrow \text{Hungarian}$
}

\footnotesize{
    ekk: $\text{Uralic} \rightarrow \text{Finnic} \rightarrow \text{Coastal Finnic} \rightarrow \text{Neva} \rightarrow \text{Central Finnic} \rightarrow \text{Estonian}$
}
\vspace{3mm}
}

We can treat each level in the path as a node and calculate the Jaccard index between the two paths. We suggest another hypothetical metric based on the number of edges from the leaf node (the concrete language) to the lowest common node (the lowest common level in the tree, which is ``Uralic'' in the above example).
However, we recognize that both methods share a significant limitation: as we can see from the example paths, the numbers of tree levels (nodes) are not evenly distributed, i.e., some language families have more fine-grained sub-levels and thus their paths contain more nodes than others. As a result, languages with shallower paths tend to be more similar to other languages than those with deeper paths, since deeper paths also mean more divergent nodes.

\subsection*{Typological similarity}

Many popular language similarity measures typically involve the comparison of typological features. One comprehensive collection of typological features is the WALS database \citep{wals}, which contains binary encodings of around 200 features for 2662 languages. Its categories include phonology (e.g., consonant and vowel inventories), lexicon (e.g. whether ``hand'' and ``arm'' are expressed differently), and word order (e.g. SVO or SOV).

URIEL is another typology database that also incorporates phylogenetic and geographic features. It sources from various typology databases: syntax features from WALS and SSWL \citep{collins2009syntactic}, phonology features from WALS and Ethnologue \citep{eberhard2019ethnologue}, and phonetic inventory features from PHOIBLE \citep{moran2014phoible}. For many features that are missing from one or more databases, URIEL infers the values based on a weighted k-nearest-neighbors algorithm with high accuracy. Alongside the database, the authors propose lang2vec, a toolkit that can be used to conveniently query all types of URIEL features.

Very recently, \citet{skirgaard2023grambank} propose Grambank, which is the largest grammatical database to date containing 195 features for 2467 languages and dialects.
Compared to other typological databases, it has a more systematic and comprehensive feature collection, which can be reflected by the association of some features to cognition and culture, e.g. politeness distinction in second person.
Another advantage of Grambank is the high coverage of its data, which has only 24\% missing values compared to, e.g., WALS' 84\% without relying on URIEL's automatic detection algorithm \citep{skirgaard2023grambank}.

\subsection*{Representational similarity}

A few recent studies use dense representations of words or languages directly to compute language similarities. \citet{conneau2019cross} integrate language embeddings into XLM, a model specialized for machine translation.
The embeddings are, however, learned during pre-training and only for pairs of languages, and therefore unrealistic to be extended to $>1000$ languages.
\citet{yu-etal-2021-language} train language embeddings from denoising autoencoders for 29 languages\footnote{\url{https://github.com/DianDYu/language_embeddings}}, which is still a small number. \citet{rama2020probing} analyze language distance based on representations from mBERT and multilingual FastText embeddings \citep{bojanowski-etal-2017-enriching}. They do so specifically by taking the averaged pairwise distances between vectors of words from a multilingual word list. Given the limited numbers of languages supported by mBERT and FastText, we note that this method is also not suitable for a large-scale comparison of language similarities.

\section{Conceptual Similarity}

Our previous work \citep{liu2023conceptualizer} proposes a pipeline for measuring language similarity based on conceptualization for 1335 languages found in the PBC. With a list of 83 concepts in total (32 from the Swadesh-100 list \citep{swadesh2017origin} and 51 derived from the Bible) and using English as the source language, we implement concept alignment using a directed bipartite graph with two steps: a forward pass and a backward pass. Specifically, for a query string in English, the forward pass is used to iteratively search for the statistically most correlated string in the target language. The backward pass is essentially the same as the forward pass, only with reversed search direction, and finds the most correlated English strings to the target language string.

To measure language similarity, we construct vectors of 100 dimensions for each of the 83 concepts and concatenate them to represent the languages. The first dimension of the vector corresponds to the concept's realization in English, and the rest dimensions represent the 99 most associated concepts. Based on such conceptual language vectors, it is possible to use a metric such as cosine similarity to quantify the conceptual relatedness between languages and form groups of conceptually similar languages. We find multiple examples of conceptual similarity that demonstrate its complementarity to geographical and genealogical closeness. For example, Plateau Malagasy, an Austronesian language spoken in Madagascar, shows similarities to both its geographically faraway Austronesian relative, Hawaiian, and Atlantic-Congo languages spoken in its neighboring countries such as Mwani and Koti. In another example, Masana, an Afro-Asiatic language spoken in Nigeria, is conceptually similar to neighboring languages Yoruba, Igbo, and Twi, despite the three being members of another language family. By analyzing the conceptual similarities of languages, we indicate that both geographical proximity and genealogical relatedness can contribute to conceptual similarity.

\section{Evaluation}

\begin{table}[h!]
\centering
\setlength\tabcolsep{7pt}
\begin{tabular}{cc||cccccc|c}
$k$ & concepts & ATLA & AUST & INDO & GUIN & OTOM & SINO & all \\
\hline\hline
\multirow{3}{*}{2} & 32 & .21  & .20  & .53  & .09  & .14  & .00 &.13\\
                    & 51 & .24  & .19  & .26  & .08  & .04  & .03 &.11\\
                    & 83 & .29  & .31  & .49  & .11  & .14  & .04 &.17\\
                    \hline 
\multirow{3}{*}{4} & 32 & .54  & .41  & .80  & .24  & .39  & .15 &.29\\
                    & 51 & .52  & .45  & .48  & .18  & .12  & .09 &.24\\
                    & 83 & .63  & .51  & .77  & .31  & .28  & .09 &.32\\
                    \hline
\multirow{3}{*}{6} & 32 & .63  & .49  & .85  & .30  & .43  & \underline{.16} &.33\\
                    & 51 & .64  & .57  & .57  & .20  & .13  & .13 &.30\\
                    & 83 & .74  & \underline{.60}  & .83  & .40  & .37  & .12 &.37\\
                    \hline
\multirow{3}{*}{8} & 32 & .68  & .53  & \textbf{.87}  & .34  & \underline{.51}  & \textbf{.18}&.36 \\
                    & 51 & .71  & .59  & .60  & .22  & .14  & .15 &.32\\
                    & 83 & \underline{.78}  & \underline{.60} & \underline{.86}  & \textbf{.42} & .36 & \textbf{.18} &\textbf{.39}\\
                    \hline
\multirow{3}{*}{10} & 32 & .73  & .56  & .84  & .34  & \textbf{.54}  & \textbf{.18} &.37\\
                    & 51 & .74  & \textbf{.61}  & .61  & .21  & .09  & .12 &.32\\
                    & 83 & \textbf{.80} & \textbf{.61}  & .83 & \underline{.41}  & .28 & \underline{.16} & \underline{.38}
\end{tabular}

\caption{\label{tab:compare_32_51_83}
Accuracy based on nearest neighbors predicted using cosine similarity between conceptual representations. Column headers from left to right: number of nearest neighbors, number of concepts (Swadesh (32), Bible (51), and All (83)), and family abbreviations (see text). \textbf{Bold} (\underline{underlined}): best (second-best) result per column. ATLA and INDO families have very high accuracy (.80 and .87), where as SINO has the lowest accuracy (.18).
}
\end{table}

\begin{table}[h!]
\centering
\setlength\tabcolsep{7pt}
\begin{tabular}{cc||rrrrrr|r}
$k$ & sim. measure & ATLA & AUST & INDO & GUIN & OTOM & SINO & all \\
\hline\hline
\multirow{4}{*}{2} & CosSim & .21 & .20  & .53  & .09  & .14  & .00 & .13 \\
                    & Hamming & .03 & .08 & .67 & .02 & .04 & .00 & .08 \\
                    & ASJP & .94 & .99 & .99 & .90 & .95 & \textcolor{teal}{1.00} & .87 \\
                    & URIEL & .98 & .99 & .92 & .84 & .97 & \textcolor{blue}{1.00} & .83 \\
                    \hline 
\multirow{4}{*}{4} & CosSim & .54  & .41  & .80  & .24  & .39  & .15 &.29 \\
                    & Hamming & \textcolor{red}{.13} & \textcolor{red}{.15} & .91 & \textcolor{red}{.05} & \textcolor{red}{.08} & \textcolor{red}{.01} & \textcolor{red}{.13} \\
                    & ASJP & .98 & \textcolor{teal}{1.00} & \textcolor{teal}{1.00} & .95 & \textcolor{teal}{.98} & \textcolor{teal}{1.00} & \textcolor{teal}{.88} \\
                    & URIEL & \textcolor{blue}{.99} & .99 & \textcolor{blue}{.96} & .99 & \textcolor{blue}{.99} & \textcolor{blue}{1.00} & \textcolor{blue}{.87} \\
                    \hline
\multirow{4}{*}{6} & CosSim & .63  & .49  & .85  & .30  & .43  & .16 &.33\\
                    & Hamming & .11 & .13 & .96 & .03 & .05 & .00 & .12 \\
                    & ASJP & .98 & \textcolor{teal}{1.00} & \textcolor{teal}{1.00} & \textcolor{teal}{.97} & \textcolor{teal}{.98} & \textcolor{teal}{1.00} & \textcolor{teal}{.88} \\
                    & URIEL & \textcolor{blue}{.99} & \textcolor{blue}{1.00} & \textcolor{blue}{.96} & \textcolor{blue}{1.00} & \textcolor{blue}{.99} & \textcolor{blue}{1.00} & .86 \\
                    \hline
\multirow{4}{*}{8} & CosSim & .68  & .53  & \textbf{.87} & \textbf{.34}  & .51 & .18 &.36 \\
                    & Hamming & \textcolor{red}{.13} & .12 & \textcolor{red}{.97} & .02 & .03 & .00 & .12 \\
                    & ASJP & .98 & \textcolor{teal}{1.00} & \textcolor{teal}{1.00} & .95 & .95 & \textcolor{teal}{1.00} & \textcolor{teal}{.88} \\
                    & URIEL & \textcolor{blue}{.99} & \textcolor{blue}{1.00} & \textcolor{blue}{.96} & \textcolor{blue}{1.00} & \textcolor{blue}{.99} & \textcolor{blue}{1.00} & .86 \\
                    \hline
                    \multirow{4}{*}{10} & CosSim & \textbf{.73}  & \textbf{.56}  & .84  & \textbf{.34}  & \textbf{.54} & .18 & \textbf{.37} \\
                    & Hamming & .11 & .10 & \textcolor{red}{.97} & .02 & .01 & .00 & .11 \\
                    & ASJP & \textcolor{teal}{.99} & \textcolor{teal}{1.00} & \textcolor{teal}{1.00} & .93 & .95 & \textcolor{teal}{1.00} & .86 \\
                    & URIEL & \textcolor{blue}{.99} & \textcolor{blue}{1.00} & \textcolor{blue}{.96} & \textcolor{blue}{1.00} & \textcolor{blue}{.99} & \textcolor{blue}{1.00} & .84 \\
\end{tabular}

\caption{\label{tab:compare_diff_metrics}
Accuracy based on nearest neighbors predicted using different similarity measures. Column headers from left to right: number of nearest neighbors, similarity or distance measure, and family abbreviations (see text). Results are calculated based on the 32 Swadesh concepts. Best result per family: \textbf{bold} (CosSim), \textcolor{red}{red} (Hamming), \textcolor{teal}{teal}: ASJP, \textcolor{blue}{blue}: URIEL. Compared to cosine similarity, Hamming distance yields very high accuracy for INDO, but low accuracy for other families. ASJP distance and URIEL similarity have comparable and good results.
}
\end{table}



\begin{table}[h!]
\centering
\begin{tabular}{cc||ccccccc|c}
$k$ & sim. & AUST & SINO & ATLA & AFRO & INDO & GUIN & OTOM & all \\
\hline\hline
2 & \multirow{5}{*}{\rotatebox{90}{Grambank}} & .75 & .58 & .78 & .52 & .61 & .09 & \underline{.42} & .48 \\
4 & & .89 & .76 & .86 & \textbf{.63} & .76 & \underline{.27} & \textbf{.58} & .61 \\
6 & & .89 & \underline{.80} & \textbf{.90} & \textbf{.63} & \underline{.78} & \textbf{.45} & \textbf{.58} & \textbf{.63} \\
8 & & \underline{.91} & \textbf{.81} & \underline{.88} & \textbf{.63} & \textbf{.82} & \textbf{.45} & \textbf{.58} & \textbf{.63} \\
10 & & \textbf{.92} & \underline{.80} & \textbf{.90} & \underline{.60} & \underline{.78} & \textbf{.45} & \underline{.42} & \underline{.62}
\end{tabular}

\caption{\label{tab:grambank_7fams}
Nearest neighbors prediction accuracy for the five largest families in Grambank and two other families used in other evaluation settings. Families are ordered based on the number of languages in Grambank. Evaluation is done using Hamming distance and Grambank representations. \textbf{Bold} (\underline{underlined}): best (second-best) result per column. Performance is very good (above 80\%) for four of the largest families Grambank. AFRO has a lower but still far above random accuracy (63\%). For the other two families (GUIN and OTOM) that have much fewer languages in Grambank, the accuracy is much lower (45\% and 58\%).
}
\end{table}

We evaluate four language similarity and distance measures. \textbf{Conceptual cosine similarity}, following \citet{liu2023conceptualizer}, calculates the similarity between two languages as the cosine similarity of their conceptual representations.
\textbf{Conceptual Hamming distance} measures the distance between two languages as the number of different elements between their binarized conceptual representations. Compared to cosine similarity, we make the hypothesis that binarized conceptual representations have the advantage of amplifying the different dimensions and making similarity more visible.
\textbf{Lexical distances based on ASJP word lists} have been calculated by \citet{ostling2023language} using mean normalized Levenshtein distance. We evaluate language similarity based on their distance matrix.
Finally, we evaluate \textbf{distance based on typological features from the URIEL database} \citep{littell2017uriel} by limiting the retrieval types to syntax, phonology, and phonetic inventories.

Following \citet{liu2023conceptualizer}, we evaluate conceptual language similarity and compare it with other similarity measures on a binary language family classification task: given a language $l$, does the majority of $l$'s $k$ nearest neighbors belong to the same family? We construct a language tree using data from Glottolog 4.7 \citep{harald_hammarstrom_2022_7398962} and consider the six top-level families with more than 50 languages in the PBC for stable results.
The six language families are: Atlantic-Congo (ATLA),
Austronesian (AUST), Indo-European (INDO), Nuclear Trans New
Guinea (GUIN), Otomanguean (OTOM, a family of indigenous
languages spoken in Mexico), and Sino-Tibetan (SINO).
We show the evaluation results in terms of classification accuracy in tables \ref{tab:compare_32_51_83} and \ref{tab:compare_diff_metrics}.

\subsection*{Conceptual cosine similarity}
In \citet{liu2023conceptualizer}, we compare language similarity by calculating cosine similarities between conceptual vectors. We evaluate vectors concatenated using three sets of concepts: 32 selected Swadesh concepts, 51 Bible concepts, and all 83 concepts. Table \ref{tab:compare_32_51_83} shows classification accuracy using different numbers of concepts. For most families, accuracy increases with the number of neighbors ($k$) until 8 and starts to drop afterward, possibly due to noise from other families. Conceptual similarity achieves good results for ATLA and INDO families (.80 and .87). INDO has the highest accuracy, which can be explained since the Conceptualizer uses English as its source language, associations to the English (thus INDO) concepts are more easily retrieved during the backward pass. For AUST, GUIN, and OTOM, conceptual similarity finds the correct family in about half of the cases. Performance for SINO languages is the worst, indicating that SINO languages, conceptually, are relatively dissimilar from each other.
These results indicate that conceptual similarity can distinguish families that are close in terms of conceptualization (e.g. INDO) from those that are not (e.g. SINO).
It can also be noted that the difference in accuracy is sometimes large between Swadesh and Bible concepts, especially for INDO and OTOM languages, an indication that the abstractness of Bible concepts contributes to variable results.

\subsection*{Conceptual Hamming distance}
In addition to cosine similarity, we use Hamming distance to measure the conceptual dissimilarity between languages. The conceptual vectors are constructed similarly, the only difference being that the dimensions are binary instead of real values, indicating the concept associated with the dimension is either found (1) or not found (0).

As shown in table \ref{tab:compare_diff_metrics}, the classification accuracy using Hamming distance is very low except for the INDO family. We hypothesize that the binarization of conceptual vectors increases the presence of English (thus INDO) concepts (see above), which causes many non-INDO languages' closest neighbors to consist of predominantly INDO languages. To test this, we investigate the distribution of the nearest neighbors in detail (Section \ref{sec:neighbor_distribution}).
We find that INDO languages constitute the majority of neighbors for all six families, which supports the assumption that conceptual Hamming distance is heavily biased toward INDO languages.

\subsection*{ASJP lexical distance}
\citet{ostling2023language} calculate lexical distances between 1012 languages in the PBC based on ASJP word lists. We use the distance matrix provided by the authors and evaluate the languages contained on the family classification task. The results are indicated by ASJP in table \ref{tab:compare_diff_metrics}. We find lexical similarity based on ASJP lists outperforms conceptual similarity with a large margin.
The classification accuracy is near 1 for all six families, suggesting that lexical similarity is a very good indication of genetic proximity.

\subsection*{URIEL typological features}
We concatenate typological vectors from the URIEL database, which include syntactic, phonological, and phonetic inventory features, to represent languages, which results in 289-dimensional binary vectors. We use the kNN-inferred features in case of missing values, and rank language similarity based on the Hamming distance. The results in table \ref{tab:compare_diff_metrics} indicate that typological features have comparable accuracy with ASJP lexical distance when it comes to language family classification.

\subsection*{Grambank typological features}
Grambank has 195 categorical features for 2467 languages in total. However, not all features are coded for every language, some of them are either not coded or unknown.
Because the features are categorical, languages must share the same subset of known features to be comparable.
For this reason, we find that using a large number of features considerably reduces the number of languages (more detailed analysis in Section \ref{sec:grambank_analysis}).
We remove unknown values and select the 50 most frequent features to represent each language. To compare the language representations, we calculate the Hamming distance.

Because Grambank has a different family distribution, we perform the evaluation for its five largest families with at least 50 languages, which are Austronesian (AUST), Sino-Tibetan (SINO), Atlantic-Congo (ATLA), Afro-Asiatic (AFRO), and Indo-European (INDO).
In addition, we provide evaluation results for GUIN and OTOM, which, however, have fewer languages in Grambank.
Table \ref{tab:grambank_7fams} shows the results for all seven families. Accuracy is overall very good (over 80\% for four of the five largest families and 63\% on average for all families). The two families with few languages in Grambank (GUIN and OTOM) have clearly worse results (45\% and 58\%).

\section{Analysis}

\subsection*{Distribution of nearest neighbors} \label{sec:neighbor_distribution}

For each of the six largest families, we investigate the average percentage of each family in the 10 nearest neighbors of their languages (results shown in \ref{tab:pct_families}). We find that in the case of conceptual Hamming distance, the neighborhoods of non-INDO languages indeed contain over 50\% INDO languages on average. This explains why prediction accuracy using Hamming distance is only good for INDO languages but very bad for other families. The percentage of same-family neighbors is also the highest for INDO when using cosine similarity, which explains why INDO is the best-performing family in this case.

The percentages for ASJP also explain the slightly worse performance of ASJP on GUIN and OTOM languages. Among the predicted closest neighbors of AUST, ATLA, and INDO languages, we find very high percentages of languages that belong to the same family, whereas for GUIN and OTOM languages, the percentages are 70\% and 76\% respectively.
Furthermore, for all six families, the percentages of languages of other families are very low (under 5\%, except for GUIN).
We notice that GUIN languages in all have many non-GUIN Papunesian neighbors, e.g. Wiru (Wiru family) and Tabaru (North Halmahera family), which share the same geographic region with them.
Given that the Papunesia region overall has a high density of language families (29 out of 120 families considered in ASJP, just under South America), we believe the high frequency of non-GUIN neighbors within a small area is one reason for the lower percentage of same-family neighbors.
OTOM languages also have a lower percentage of OTOM neighbors on average. However, the non-OTOM neighbors are more varied compared to GUIN's case, although we do encounter many South American (geographically close to where OTOM languages are used) neighbors.
We also suppose that for both GUIN and OTOM families, conceptualizations may be distinct from each other, as reflected by their lower classification accuracy in Table \ref{tab:compare_32_51_83}.

Similarly, percentages of same-family languages are slightly lower for INDO and GUIN when using URIEL typological features (both 88\%), which explains why accuracy on the two families is slightly worse than other families.

For Grambank, we additionally evaluate the predicted neighbors of AFRO, one of its largest families (Table \ref{tab:pct_families_grambank}). All of the five largest families in Grambank have a clear majority of same-family neighbors. Among the five families, AUST has the highest and AFRO the lowest percentage of same-family neighbors. This is consistent with the results in Table \ref{tab:grambank_7fams}.
The two less represented families in Grambank (GUIN and OTOM) have much smaller proportions of same-family neighbors, which is also consistent with their lower accuracy.

\subsection*{WALS features}

It is known that WALS without automatic feature detection (e.g. as done by \citet{littell2017uriel}) has many missing values \citep{skirgaard2023grambank}.
We make the hypothesis that contributing linguists who specialize in different language families may restrain themselves to a limited range of features.
This means the concentration of features for specific families may vary, affecting the comparability across families and potentially making genealogical prediction easier.

To examine it, we calculate the coverage of WALS features (syntax and phonology) for the six largest families.
The coverage of most features for INDO and SINO languages tends to be higher than others. For other families, the coverage is more variable.
Some features have very low coverage for all six families (up to around 10\%), e.g. ``ergative-absolutive mark''. Others have higher coverage for some families. For example, ``polar question word'' has 40\% coverage for INDO and SINO and around 20\% for other families.
A small number of features are lacking for entire families, e.g. features related to the oblique position are missing for all GUIN languages.

\subsection*{Grambank features} \label{sec:grambank_analysis}
Grambank has systematic feature encodings for its 2467 languages, most of which have close to all 195 feature entries in the database. In practice, however, we find that many features have the ``unknown'' value, meaning the entry cannot be used for language similarity comparison, and that few values are shared by a large number of languages.
\citet{lesage-etal-2022-overlooked} mention this by stating that the ``description level'' of Grambank features (i.e. how well they are documented across different languages) varies strongly.

\begin{figure}[h]
\centering
\includegraphics[width=\textwidth]{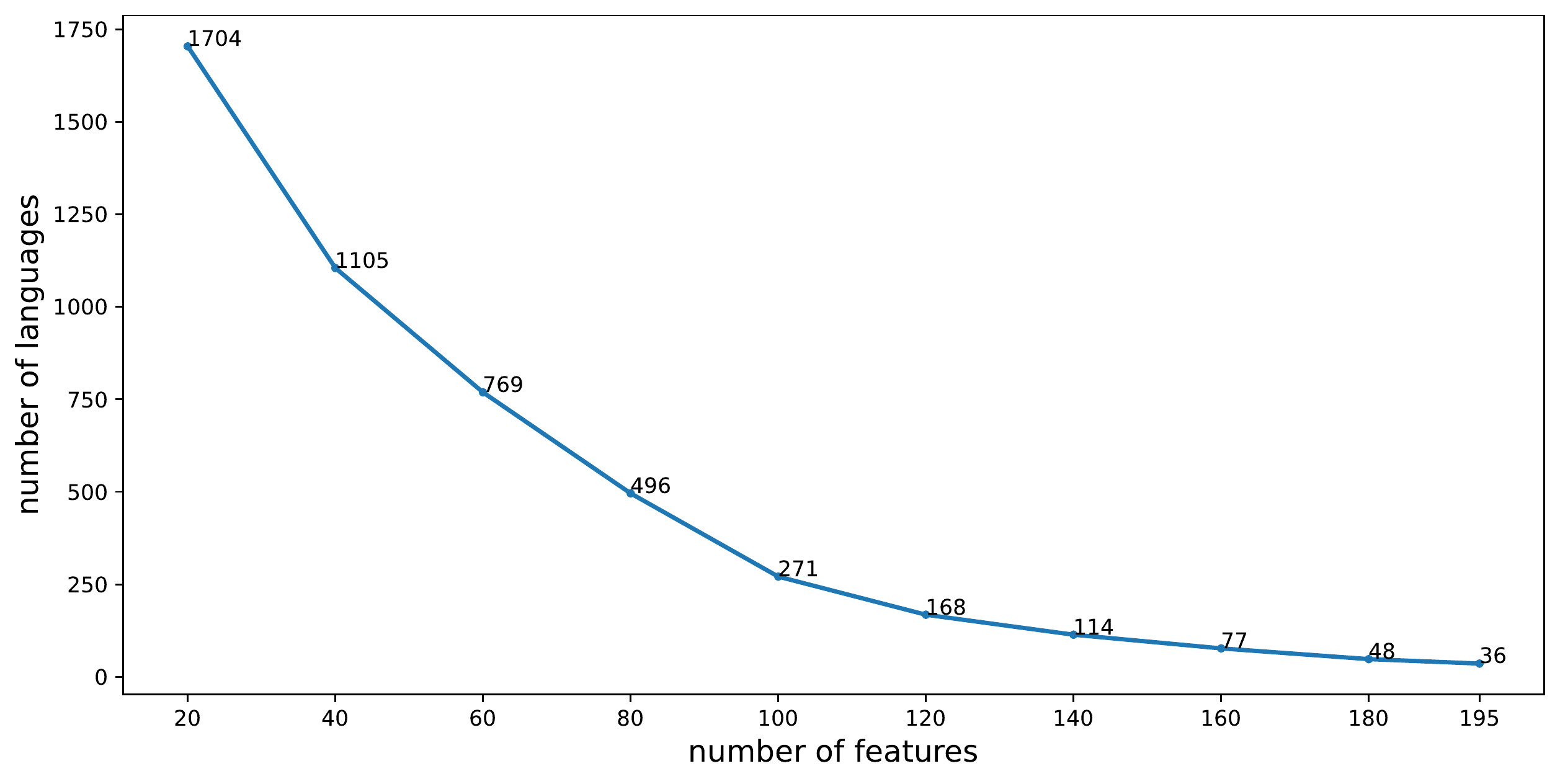}
\caption{This graph demonstrates the trade-off between the number of Grambank features to include, i.e. the dimensionality of language representations, and the number of languages that fulfill the set of features.}
\label{fig:grambank_feature_counts}
\end{figure}

Figure \ref{fig:grambank_feature_counts} shows the relationship between the number of most frequent features and the number of languages that share those features. It can be easily observed that the number of languages available for comparison decreases rapidly when we include a larger number of features.
For example, while 1105 of the 2467 languages can be compared using 40 features, expanding the size of the feature set to 100 more than quarters the number of comparable languages.

We investigate the coverage of Grambank features for the five largest families. Although the features tend to be more evenly distributed across the five families, many still have much lower coverage for one family than the others.
For example, feature \textit{GB325: Is there a count/mass distinction in interrogative quantifiers?} is missing for half of the AFRO languages, but covers over 80\% INDO languages.

\section{Conclusion}\label{sec:conclusion}

As far as we know, our work is the first to perform an empirical evaluation of different language representations for their predictive performance of genealogical language similarity.
Our evaluation covers a recently proposed work on conceptual language similarity \citep{liu2023conceptualizer} and a newly released grammatical database, Grambank \citep{skirgaard2023grambank}, including its first implementation on language similarity prediction and an analysis of its drawbacks.
We have previously shown that conceptual similarity has interesting complementarities to existing similarity measures.
For example, languages that would otherwise not be considered similar within the genetic or typological frameworks, such as Tagalog and Spanish, demonstrate similarities on the conceptual level.
However, as shown by the evaluation results (Table \ref{tab:compare_diff_metrics}), conceptual similarity is weaker than lexical and typological similarities when it comes to accuracy in genetic similarity classification.
This is not surprising, as lexical and typological features tend to be similar to languages of the same family.
Specifically, many typological features, e.g. from WALS, are related to word order or lexicon, which are likely to be influenced by assimilation or geographic proximity.
This in turn often indicates genealogical proximity between languages.
We believe that there is currently a lack of evaluation tasks that specifically suit the conceptual view of language similarity.
Therefore, if the main objective is classification accuracy, typology or lexical features offer a strong signal for genetic relatedness.
However, for someone interested in a high-level comparison of languages beyond the language family boundary, conceptual similarity is likely to offer valuable insights.

Given the presence of more Indo-European languages in the nearest neighbors overall, our next step would be to evaluate whether and how the source language influences the results.

\section*{Declarations}

\textbf{Competing Interests} Authors are required to disclose financial or non-financial interests that are directly or indirectly related to the work submitted for publication. \url{https://www.springer.com/journal/10579/submission-guidelines} - ``Competing Interests''

\begin{appendices}

\section{Percentages of families in predicted nearest neighbors} \label{app:pct_families}

\begin{table}[h!]
\centering
\begin{tabular}{c|c|cccccc}
\multirow{2}{*}{sim. measure} & \multirow{2}{*}{src language} & \multicolumn{6}{c}{\% predicted neighbors} \\
 &  & ATLA & AUST & INDO & GUIN & OTOM & SINO \\
\hline\hline
\multirow{6}{*}{CosSim} & ATLA & 41 & 7 & 13 & 1 & 4 & 5 \\
 & AUST & 19 & 31 & 9 & 2 & 3 & 5 \\
 & INDO & 13 & 3 & 55 & 0 & 1 & 2 \\
 & GUIN & 15 & 17 & 2 & 18 & 3 & 4 \\
 & OTOM & 18 & 5 & 2 & 1 & 24 & 3 \\
 & SINO & 19 & 9 & 21 & 1 & 1 & 14 \\
\hline
\multirow{6}{*}{Hamming} & ATLA & 20 & 4 & 56 & 0 & 0 & 1 \\
 & AUST & 12 & 14 & 52 & 0 & 0 & 1 \\
 & INDO & 8 & 3 & 69 & 0 & 0 & 0 \\
 & GUIN & 13 & 6 & 52 & 5 & 0 & 3 \\
 & OTOM & 15 & 1 & 51 & 0 & 5 & 1 \\
 & SINO & 10 & 6 & 58 & 0 & 0 & 2 \\
\hline
\multirow{6}{*}{ASJP} & ATLA & 89 & 2 & 0 & 1 & 0 & 1 \\
 & AUST & 0 & 99 & 0 & 0 & 0 & 0 \\
 & INDO & 0 & 0 & 99 & 0 & 0 & 0 \\
 & GUIN & 7 & 7 & 1 & 70 & 0 & 1 \\
 & OTOM & 4 & 5 & 1 & 3 & 76 & 2 \\
 & SINO & 1 & 1 & 0 & 0 & 0 & 97 \\
\hline
\multirow{6}{*}{URIEL} & ATLA & 96 & 0 & 0 & 0 & 0 & 0 \\
 & AUST & 0 & 99 & 0 & 0 & 0 & 0 \\
 & INDO & 0 & 2 & 88 & 0 & 0 & 0 \\
 & GUIN & 0 & 0 & 0 & 88 & 0 & 0 \\
 & OTOM & 0 & 1 & 0 & 0 & 98 & 0 \\
 & SINO & 0 & 1 & 0 & 0 & 0 & 96
\end{tabular}
\caption{\label{tab:pct_families}
Src language: source language for which the closest neighbors are predicted; \% predicted neighbors: averaged percentages of languages belonging to each family among the ten nearest neighbors. All families have a majority of INDO neighbors when using Hamming distance. A lower \% of same-family neighbors is possibly correlated to a lower accuracy (details in Section \ref{sec:neighbor_distribution}).
}
\end{table}

\begin{table}[h!]
\centering
\begin{tabular}{c|c|ccccccc}
\multirow{2}{*}{sim.} & \multirow{2}{*}{src language} & \multicolumn{7}{c}{\% predicted neighbors} \\
 &  & AUST & SINO & ATLA & AFRO & INDO & GUIN & OTOM \\
\hline\hline
\multirow{7}{*}{\rotatebox{90}{Grambank}} & AUST & 79 & 2 & 2 & 1 & 0 & 0 & 1 \\
 & SINO & 3 & 62 & 0 & 0 & 1 & 2 & 0 \\
 & ATLA & 7 & 1 & 76 & 5 & 2 & 0 & 0 \\
 & AFRO & 6 & 1 & 15 & 47 & 7 & 0 & 1 \\
 & INDO & 3 & 3 & 2 & 3 & 64 & 0 & 0 \\
 & GUIN & 0 & 17 & 0 & 0 & 0 & 24 & 0 \\
 & OTOM & 45 & 0 & 0 & 0 & 0 & 0 & 29
\end{tabular}
\caption{\label{tab:pct_families_grambank}
Percentage of predicted families using Grambank representations. Src language: source language for which the closest neighbors are predicted; \% predicted neighbors: averaged percentages of languages belonging to each family among the ten nearest neighbors. All five families with the most languages in Grambank have majority same-family neighbors. The two less represented families (GUIN and OTOM) have much higher proportions of neighbors from other families, which is consistent with their lower accuracy (Table \ref{tab:grambank_7fams}).
}
\end{table}

\end{appendices}


\bibliography{sn-bibliography}


\end{document}